\documentclass[runningheads]{llncs}
\usepackage[T1]{fontenc}
\usepackage{amsmath,amssymb,amsfonts}
\usepackage{algorithmic}
\usepackage{graphicx}
\usepackage{textcomp}
\usepackage{xcolor}
\usepackage{hyperref}
\usepackage{graphicx}
\begin{document}
\title{Merging Multiple Datasets for Improved Appearance-Based Gaze Estimation}
%

\author{Liang Wu\orcidID{0000-0001-5214-7715} \and
Bertram E. Shi\orcidID{0000-0001-9167-7495} }
%
%
\institute{Hong Kong University of Science and Technology, Hong Kong 
\email{liang.wu@connect.ust.hk}\\
\email{eebert@ust.hk}}

\maketitle              
\begin{abstract}
Multiple datasets have been created for training and testing appearance-based gaze estimators. Intuitively, more data should lead to better performance. However, combining datasets to train a single estimator rarely improves gaze estimation performance. One reason may be differences in the experimental protocols used to obtain the gaze samples, resulting in differences in the distributions of head poses, gaze angles, illumination, etc. Another reason may be the inconsistency between methods used to define gaze angles (label mismatch). We propose two innovations to improve the performance of gaze estimation by leveraging multiple datasets, a change in the estimator architecture and the introduction of a gaze adaptation module. Most state-of-the-art estimators merge information extracted from images of the two eyes and the entire face either in parallel or combine information from the eyes first then with the face. Our proposed \textit{Two-stage Transformer-based Gaze-feature Fusion} (TTGF) method uses transformers to merge information from each eye and the face separately and then merge across the two eyes. We argue that this improves head pose invariance since changes in head pose affect left and right eye images in different ways. Our proposed \textit{Gaze Adaptation Module} (GAM) method handles annotation inconsistency by applying a Gaze Adaption Module for each dataset to correct gaze estimates from a single shared estimator. This enables us to combine information across datasets despite differences in labeling. Our experiments show that these innovations improve gaze estimation performance over the SOTA both individually and collectively (by 10\% -  20\%). Our code is available at \url{https://github.com/HKUST-NISL/GazeSetMerge}.

\end{abstract}

\section{Introduction}

Estimation of human gaze plays important roles in many applications, such as human-computer interaction \cite{chen2019using,pi2017probabilistic}, virtual reality \cite{patney2016towards}, attention analysis \cite{recasens2015they,chong2020detecting} and psychological studies \cite{gehrer2018implementing}. 

Conventional methods, such as those based on pupil center corneal reflections (PCCR), use 3D eye models to compute the gaze direction \cite{guestrin2006general}. These require special measurement setups, such as active infrared illumination, to estimate model geometry. In contrast, appearance-based gaze estimators use input from commonly available RGB web cameras, which are more convenient and less expensive. Unfortunately, estimates from them are less accurate than those from PCCR-based systems. The current lowest reported within-person error of gaze estimation is $4.04^{\circ}$ \cite{o2022self} on MPIIFaceGaze. In contrast, manufacturers of PCCR-based systems typically report accuracies of less than one degree. 

However, the gap between the two continues to shrink, most recently due to the use of Convolutional Neural Networks (CNN) \cite{zhang2015appearance,zhang2017s,chen2018appearance} and transformers. Many CNN architectures have been proposed for appearance-based gaze estimation. Zhang et al. employed a multi-modal model that used eye images and an estimated head pose vector as inputs to estimate gaze direction \cite{zhang2015appearance}. Later, they applied spatial weighting to feature maps from the face image to enhance information from eye regions \cite{zhang2017s}. Other studies used three separate pipelines to extract features from images of the head and the two eyes and then fused them to predict the gaze \cite{chen2018appearance,krafka2016eye}. Merging information from the eyes and the face improves estimation accuracy. 

Since appearance-based gaze estimators rely heavily on training data, many datasets have been proposed to train gaze estimators. Initial datasets were collected under fairly well-controlled and limited conditions (e.g., ranges of head poses and gaze angles). More recent datasets have been collected on conditions of increased diversity. The availability of more data can potentially increase the performance of appearance-based gaze estimators, but can also introduce new challenges. This paper seeks to address two of these challenges. 

First, increases in the head pose range have spurred the development of new architectures that combine information from images of the two eye regions (which primarily indicate gaze direction in head-centric coordinates) and an image of the entire face (which primarily indicates head pose). Many SOTA (state-of-the-art) methods combine this information in parallel \cite{chen2018appearance}, or combine information from the eyes first followed by the face image \cite{krafka2016eye}. 

To improve upon these approaches, we propose a Two-stage Transformer-based Gaze-feature Fusion (TTGF) architecture, which combines information from each eye image with the face image separately and then integrates information across the two eyes. This approach is motivated by the fact that the head-centric gaze directions of the two eyes differ and should thus each be merged with the face image. This may also compensate for situations where the reliability of information from the two eyes may differ, e.g., due to occlusion. 

Second, although intuitively increasing the amount of data by combining datasets should improve performance, inconsistencies in annotation among datasets make it difficult to improve accuracy by simply combining multiple gaze datasets. To provide a normalized gaze annotation, a common scheme is to rotate the gaze vector from the gaze origin to the target point by a rotation matrix that depends upon the head pose \cite{zhang2018revisiting}. Differences between the methods for head pose estimation and target point estimation lead to inconsistency among different datasets. Even when the subject's head is constrained by a chin rest \cite{smith2013gaze}, head pose estimation error can still exist due to the placement of the subject's head in the chin rest. 

To address this, we propose the use of a Gaze Adaption Module (GAMs) for each dataset, which adjusts the gaze label from a shared estimator so it is consistent with the dataset of the source image. This enables multi-dataset training by simply adding GAM to the model's gaze regression head. 
 
Our experimental results demonstrate that these two innovations lead to state-of-the-art performance on multiple datasets, under training with both single datasets and mixed datasets.

\section{Related Work}
\noindent 
\textbf{Gaze Estimation Methods}\quad Gaze estimation methods can typically be categorized as either model-based or appearance-based. Model-based methods usually construct the 3D model of the head and eyes. The gaze direction is calculated by utilizing geometric information \cite{chen20083d,valenti2011combining,wood2014eyetab,guestrin2006general}. Model-based methods usually require time-consuming personal calibration to fit the subject-specific parameters, such as cornea radius and kappa angles. 

In contrast, appearance-based methods directly learn mapping functions from a large number of image-gaze sample pairs. 
Early approaches used conventional regression to perform the mapping \cite{tan2002appearance,lu2014adaptive,williams2006sparse}. More recently, CNNs have significantly improved the performance of appearance-based gaze estimation.  Zhang et al. proposed the first CNN-based network to regress the gaze direction from a cropped eye image, and a head pose vector \cite{zhang2015appearance}. They later proposed to use the learnable spatial weights to enhance the information from the eye regions in the face image \cite{zhang2017s}. Krafka et al. proposed iTracker, a multi-region CNN model, which takes both the head and eye images as input. To further improve the accuracy, Chen et al. investigated the dilated convolution layers to efficiently increase the receptive field sizes of the features \cite{chen2018appearance}. Researchers have now started to use transformer-based networks, which can further improve gaze estimation accuracy \cite{cheng2021gaze,tu2022end,cai2021gaze}. 

\noindent 
\textbf{Transformers}\quad The Transformer architecture was first introduced by Vaswani et al. for natural language processing \cite{vaswani2017attention}. It consists of self-attention layers, layer normalization, and multi-layer perceptron layers. Compared with recurrent networks, the global computations and efficient memory of self-attention layers make transformers more suitable for long sequences. 

The Vision Transformer (ViT) was proposed by Dosovitskiy et al. for image classification tasks \cite{dosovitskiy2020image}. ViT divides one image into non-overlapping patches. A transformer encoder is applied to the features extracted from the patches. Transformers have achieved state-of-the-art in large-scale image classification tasks, leading to their application to many other vision tasks \cite{xie2021segformer,kim2021hotr,zhang2022transformer}.

Recently, a few researchers have explored the capability of transformers in gaze estimation. Cheng et al. proposed GazeTR-Hybrid where they used convolutional neural networks to extract the feature map of an input head image, then treated the features at different positions as a sequence of features input to a transformer encoder \cite{cheng2021gaze}. Cai et al. proposed iTracker-MHSH \cite{cai2021gaze}. Inspired by iTracker, it uses a transformer to integrate the features of the head and eye images. 

\noindent
\textbf{Mixed Dataset Training}\quad  There are two main advantages to mixed dataset training. First, it provides a single model applicable to multiple datasets. Second, model training may benefit from the increased amount of data. Mixed dataset training has been applied to many computer vision tasks, such as person reidentification \cite{lv2018unsupervised,li2019cross}, monocular depth estimation \cite{ranftl2020towards}, semantic image segmentation \cite{DBLP:conf/aaai/HeZZT20,lambert2020mseg}, video quality assessment \cite{li2021unified,korhonen2019two} and 3D object detection \cite{zhang2021srdan}. Addressing the challenges of mixed dataset training is task-specific. For example, to mix image segmentation datasets, category merging needs were conducted before training \cite{DBLP:conf/aaai/HeZZT20,lambert2020mseg}. For video quality assessment \cite{li2021unified}, the challenge was to resolve inconsistent ranges of subjective quality scores across datasets.

To the best of our knowledge, we are the first to propose mixed dataset training for gaze estimation. There are two challenges that must be addressed. First, the distribution of gaze vectors and head poses varies between different gaze datasets. Second, there exists annotation inconsistency in gaze vectors from different gaze datasets. 

\section{Annotation Inconsistency}

The gaze vector is defined as the vector starting from the gaze origin to the gaze target. Gaze dataset collection requires an experimental setup to capture three types of information in camera coordinates: 1) the position of the visual target $P_t$, 2) the position of gaze origin $P_o$, and 3) the head pose $R$ \cite{rodrigues2010camera}. However, different datasets utilize different methods to get these values, leading to different annotations.

\noindent 
\textbf{Inconsistency in gaze target estimation}\quad Usually, the visual target is indicated by a moving dot on a screens. To determine the position of the dot target, the intrinsic parameters of the camera must be obtained beforehand. MPIIGaze uses a mirror-based calibration method \cite{rodrigues2010camera} to estimate the 3D positions of each screen plane. Finally, the position of the moving dot is computed based on the screen size and resolution. In addition to a moving dot on the screen, EYEDIAP has an additional floating ball visual target. Its position is estimated first in an RGB-D sensor coordinate system and then transformed to the camera coordinate system. Imprecision in the RGB-D sensor, errors in the screen-to-camera calibration and RGB-D to-camera calibration will all contribute to the inconsistency of the gaze target position $p_t$.

\noindent 
\textbf{Inconsistency in gaze origin and head pose estimation}\quad There are inconsistencies between datasets in the selection of gaze origin and the estimation of head pose. In early work, gaze was estimated eye images, where the eye center defined the gaze origin \cite{zhang2015appearance,park2018deep,liu2019differential}. More recently, people estimate gaze from the whole head image, where the gaze origin is usually set at the center of the head \cite{zhang2020eth,fischer2018rt,funes2014eyediap}. To get the 3D head pose, MPIIGaze and ETH-XGaze detect landmarks from the 2D head image and fit a 3D morphable model of the head to the detected landmarks. EYEDIAP directly uses the depth data from the RGB-D sensor to fit a 3D Morphable Model.

\section{Method}

\label{sec:method}

\begin{figure*}[t]
  \vspace{0.5cm}
  \centering
  \includegraphics[width=0.95\textwidth]{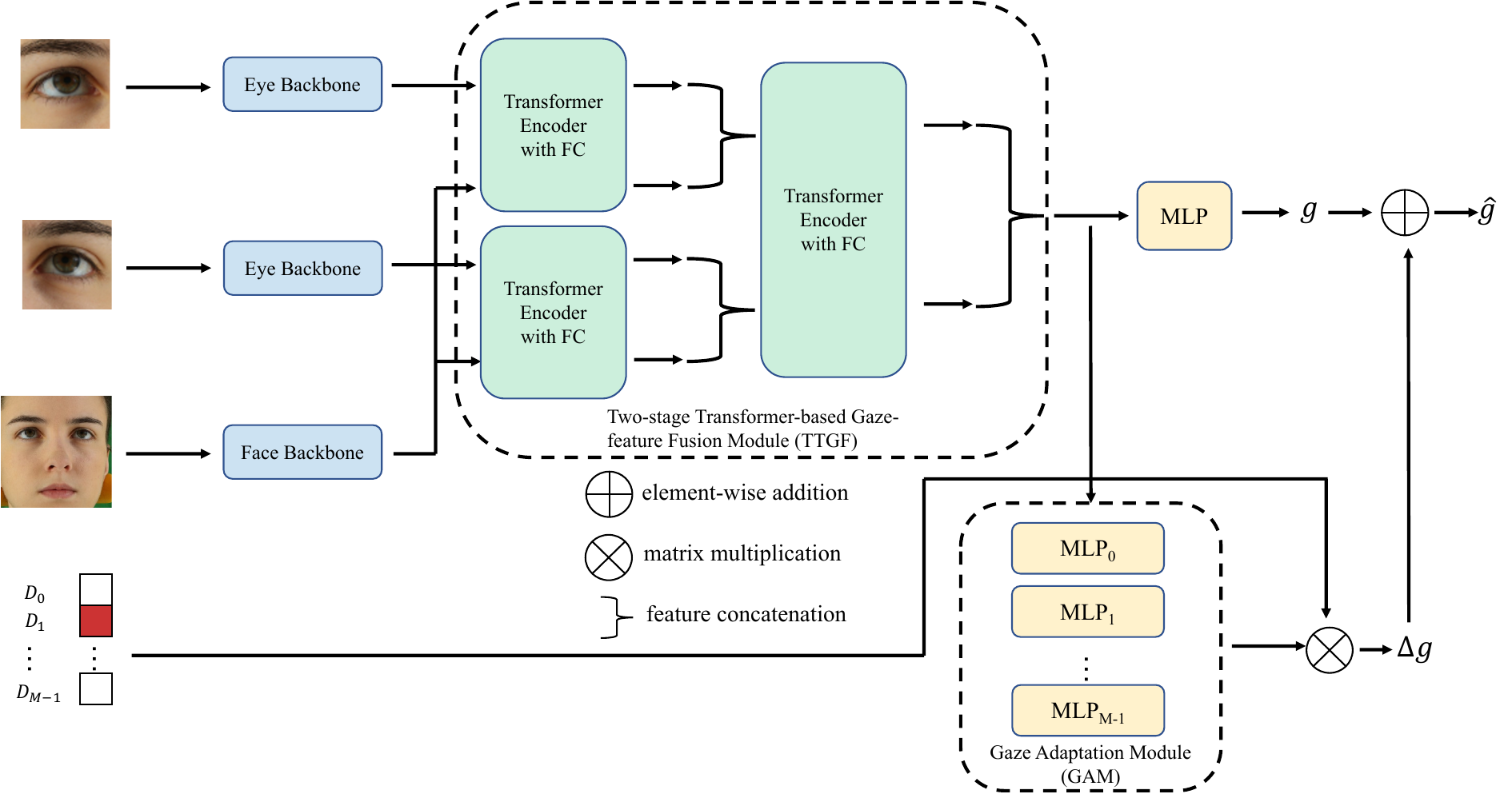}
  \caption{The proposed framework contains two modules: 1) TTGF and 2) GAM. The TTGF applies two-stage feature fusion to the features of the head and eyes with transformers, and the GAM produces a gaze offset to adjust the predicted gaze for mixed datasets training.}
  \label{fig:framework}
\end{figure*}

Fig.\ref{fig:framework} shows our framework, which consists of an eye-head transformer-based feature fusion module for gaze estimation followed by a set of gaze adaptation modules. We described these in more detail below.

\subsection{Feature Fusion with Transformers}

\label{sec:model}
A typical transformer encoder contains $L$ transformer blocks, each containing multi-head self-attention (MHSA) layers, layer normalization (LN), and multi-layer perceptron layers (MLP). To process an input feature matrix $\boldsymbol{Z}\in\mathbb{R}^{n\times d}$, MHSA projects $\boldsymbol{Z}$ into $\boldsymbol{Q} \in \mathbb{R}^{n\times d_k}$, keys $\boldsymbol{K} \in \mathbb{R}^{n \times d_k}$ and values $\boldsymbol{V} \in \mathbb{R}^{n\times d_v}$ where $n$ is the number of tokens and $d, d_k, d_v$ are the dimension of the feature, key/query and value. 

The attention is computed through the following equation:
\begin{equation}
    \text{Attention}(\boldsymbol{Q}; \boldsymbol{K}; \boldsymbol{V}) = \text{softmax}(\frac{\boldsymbol{Q}\boldsymbol{K}^T}{\sqrt{d_k}}\boldsymbol{V}).
\end{equation}
Combined with LN and MLP, the overall equations for the transformer encoder with $L$ transformer blocks are

\begin{align}
    z'_l &= \text{MSA}(\text{LN}(z_{l-1})) + z_{l-1}, &l=1...L,\\
    z_l &= \text{MLP}(\text{LN}(z'_{l})) + z'_{l}, &l=1...L,\\
    y &= \text{LN}(z_L).
\end{align}
 
Krafka et al. proposed iTracker \cite{krafka2016eye} to estimate gaze by integrating the features of the head and eyes using several fully connected layers. To better fuse features, we propose the two-stage transformer-based gaze-feature fusion (TTGF) architecture shown in Fig. \ref{fig:framework}. 
This architecture applies three transformer encoders to fuse the features from the head and eye images in two fusion steps, 1) head-eye fusion and 2) left-right fusion. The idea of using two-step fusion is based on the intuition combining information of the head and one eye enable rough inference of the person's gaze direction. The second step combines the two rough estimates into a single more precise estimate.


In our design, the architectures of all three fusion modules are identical. One TGF module accepts two gaze-related features and produces a fused feature. We describe the computation in ta TGF formally with the following equation:
\begin{equation}
    \text{TGF}(f^\ast, f^\dag) = \text{CAT}(\text{FC}(\text{Trans}([f^\ast; f^\dag]))), 
\end{equation}
where $\text{Trans}([f^\ast; f^\dag])$ is the transformer used for fusing the head-eye features or eye-eye features, $\text{FC}$ is a linear layer used to project the features to a specific size, and $\text{CAT}$ concatenates the outputs of the transformer to generate fused features. In the head-eye fusion stage, each eye feature $f^{le}$ or $f^{re}$ is fused with the head feature $f^h$:

\begin{equation}
    f^{lh} = \text{TGF}^{lh}(f^{le}, f^{h})
\end{equation}
\begin{equation}
    f^{rh} = \text{TGF}^{rh}(f^{re}, f^{h})
\end{equation}
In the second stage, the two fused eye-head features are fed into a third TGF module to fuse features from left and right:

\begin{equation}
    f^{lr} = \text{TGF}^{lr}(f^{lh}, f^{rh})
\end{equation}
Finally, the fused feature $f^{lr}$ is fed to an MLP to get the predicted gaze $g$:

\begin{equation}
    g = \text{MLP}(f^{lr}).
\end{equation}

\subsection{Gaze Adaptation Module}
\label{sec:gam}
Suppose we have $M$ gaze datasets, $D = \{D_0, D_2, ..., D_{M-1}\}$. Typically, we need to train $M$ models: one for each dataset to get good performance. A model trained on $D_i$ typically performs poorly on $D_j$ where $i \neq j$. 

Instead, our approach trains only one model and $M-1$ Gaze Adaptation Modules (GAMs). The GAM is a module consisting of a $M$ MLPs, one for each dataset $i \in \{0,\ldots,M-1\}$. Each MLP, $\text{MLP}_i(\cdot)$, accepts the extracted feature $f^{lr}$ and produces a gaze offset assuming the sample comes from dataset $i$. $D_0$ is regarded as the anchor dataset, so its offset is always zero, i.e., $\text{MLP}_0(\cdot) =\boldsymbol{0}$ and does not need to be trained. For the others, the MLP has two layers with GELU nonlinearities. If the sample comes from dataset $i$, the corrected gaze vector is given by $\hat{g} = g + {\Delta{g}}$, where $\Delta{g} = \text{MLP}_i(f^{lr})$. 

\subsection{Architecture Details}
\label{sec:impl}

The whole architecture contains three pipelines for the face and two eye images. All the backbones are ResNet18 networks, which are initialized from the model trained on ImageNet. The input face image size is $224 \times 224 \times 3$. We crop the eye patches according to the landmarks and use RoI align to resize the cropped patches to $128\times 128 \times 3$. The estimated gaze contains the yaw and pitch representing the 3D gaze direction in the camera coordinate system. We chose L1 loss as the loss function for gaze estimation.

For TTGF, we set the number of heads of all MSAs as 8 and the hidden size of the MLP is 2048. We use 8 repeated blocks in each transformer encoder. After each transformer encoder, the features are projected with a linear layer whose output size is 128. For the MLPs for both gaze regression and the GAMs, the sizes of the hidden layers are identically set to 128.


\section{Experiments}
\label{sec:exp}
In this section, we introduce the experimental settings and the evaluation datasets we selected and evaluate our proposed TTGF and GAM in two types of experiments. We first compare our method with the state-of-the-art methods for gaze estimation performance. Then we perform ablation studies to determine the effects due to TTGF and GAM respectively and study the effect of multiple dataset training.  

\noindent 
\textbf{Dataset for evaluation}\quad For evaluating gaze estimation performance, we used three gaze datasets to evaluate the gaze estimation performance as shown in Table \ref{tab:data}: MPIIFaceGaze \cite{zhang2017s}, RT-GENE \cite{fischer2018rt}, and EYEDIAP \cite{funes2014eyediap}. MPIIFaceGaze dataset is based on MPIIGaze, but includes face and eye images.  It contains 45K images collected from 15 subjects. We used leave-one-person-out cross-validation with this dataset. The RT-GENE dataset consists of 123K samples from 15 participants. We used three-fold cross-validation with this dataset. The raw data of the EYEDIAP dataset has 94 videos collected from 16 subjects. We used the sampling scheme from \cite{zhang2020eth} to extract face images and four-fold cross-validation. For our experiments on multi-dataset training, we trained 15 models (one for each subject left out from MPIIFaceGaze), where each person was assigned to one of the folds in the other two datasets. Performance for each fold in the other two datasets was computed by averaging the performance of the models from the MPIIFaceGaze subjects assigned to that fold.

\begin{table*}[ht]
    \centering
    \caption{Overview of the datasets used for evaluation and anchor dataset in our experiments. We show the number of subjects, the range of gaze, and the head pose in both horizontal and vertical directions in the camera coordinate systems.}
    \begin{tabular}{l|c|c|c|c}
    \hline
    Dataset & \# Subjects & Gaze & Head Pose & \# Data \\ \hline
    MPIIFaceGaze \cite{zhang2017s} & 15 & $\pm20$\textdegree, $\pm20$\textdegree & $\pm15$\textdegree, $30$\textdegree & 45K images \\ 
    RT-GENE \cite{fischer2018rt} & 15 & $\pm40$\textdegree, $-40$\textdegree & $\pm40$\textdegree, $\pm40$\textdegree & 123K images \\ 
    EYEDIAP  \cite{funes2014eyediap} & 16 & $\pm25$\textdegree, $20$\textdegree & $\pm15$\textdegree, 30\textdegree & 94 videos \\ 
    ETH-XGaze  \cite{zhang2020eth} & 110 & $\pm120$\textdegree, $\pm70$\textdegree & $\pm80$\textdegree, $\pm80$\textdegree & 1.1M images \\ \hline
    \end{tabular}
    \label{tab:data}
\end{table*}

\noindent 
\textbf{Anchor dataset and pre-training}\quad ETH-XGaze \cite{zhang2020eth} is a large-scale gaze dataset that consists of 1,083,492 image samples from 110 participants (47 female and 63 male). It has the largest range of head poses compared to the evaluation dataset and the gaze direction is evenly sampled both horizontally and vertically as shown in Table \ref{tab:data}. The large variation and scale make it a suitable dataset as the anchor dataset $D_0$ and for pre-training. The whole dataset contains three parts: the training set, the within-dataset, and the person-specific evaluation set. The training set has 765K images of 80 subjects. We use this part as the anchor set and also for pre-training. The person-specific evaluation consists of 15 subjects but is not related to this task. The within-dataset which includes 15 subjects is used for validation of multiple datasets training and the pre-training model. 

\noindent 
\textbf{Experimental settings}\quad The optimizer applied for model training is AdamW with a linear scheduled warm-up strategy. The initial learning rate is set to 0.0001 for all the training and uses the exponential schedule to update it. For multiple-set training, in each iteration, we randomly sample the same number of samples from each set to form a batch fed to the model. The batch size is set to 64. The number of iterations in one training epoch is determined by the size of the dataset with the smallest number of samples. The number of epochs is 50 and gamma is 0.96. For single-set training for the TTGF-only model, the batch size is also set to 64. For ETH-XGaze, we train the model for 50 epochs with the exponential gamma setting to 0.95. For MPIIFaceGaze and RT-GENe, the total number of epochs is 30 epochs with the exponential gamma setting to 0.95. For EYEDIAP, the number of epochs is 50 and gamma is 0.096. Our experiments are all conducted on a single GeForce RTX 3090 GPU. 

\subsection{Comparison with state-of-the-art methods}
In this part, we compare the gaze estimation performance of our proposed model with state-of-the-art methods. Our model is a single model trained on multiple datasets: one anchor dataset and three evaluation datasets, while the existing methods were tested with separated models for different evaluation datasets. We trained our TTGF-only model on ETH-XGaze and got a testing error of 3.58\textdegree and the proposed TTGF+GAM trained on multiple datasets achieved a slightly better error of 3.54\textdegree. 

\begin{figure*}[ht]
  \centering
  \includegraphics[width=0.95\textwidth]{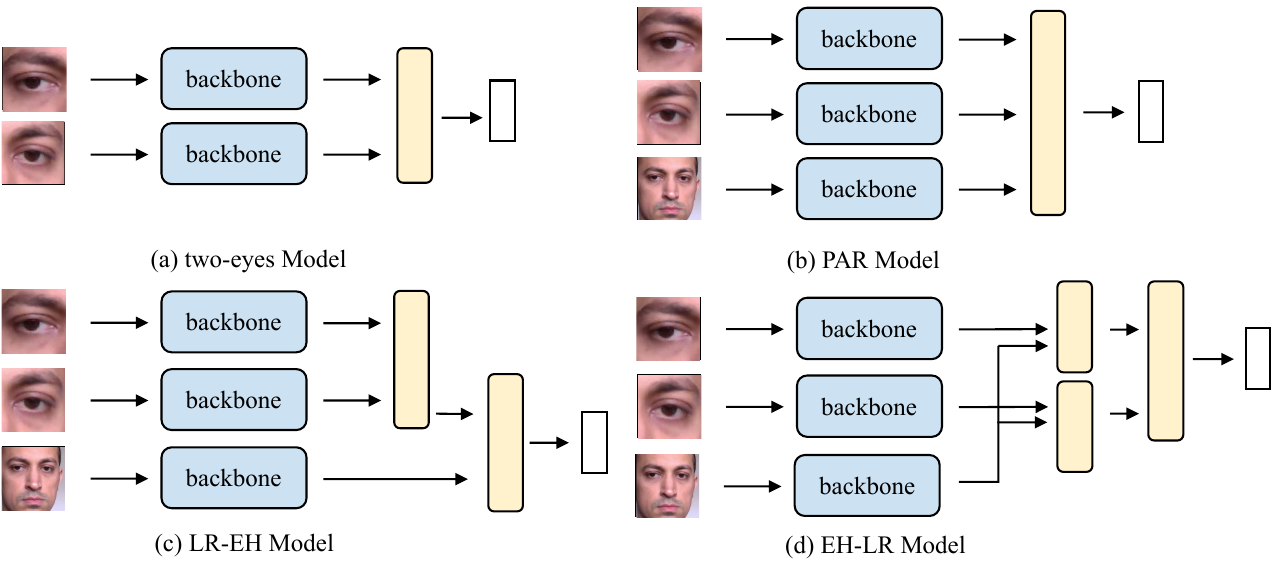}
  \caption{Four types of feature fusion for gaze estimation models: (a) two-eyes model uses the cropped eye patches as inputs. (b) PAR indicates left eye, right eye, and head features are combined in parallel. (c) LR-EH indicates that left and right eye features are combined first then combined head features. (d) EH-LR indicates that single eye and head features are combined first followed by a combination across the left
and right..}
  \label{fig:types}
\end{figure*}

\begin{table*}
    \centering
    \caption{Comparison with the state-of-the-art methods. The proposed method outperforms state-of-the-art results in estimation error.} 
    \begin{tabular}{l|c|c|c|c|c}
    \hline
    Model          &Transformer&Feature Fusion& MPIIFaceGaze & RT-GENE & EYEDIAP  \\ \hline
    FullFace \cite{zhang2017s}&NO&Face Only&4.93\textdegree&10.00\textdegree&6.53\textdegree\\ \hline
    RT-GENE \cite{fischer2018rt}&NO&Two Eyes&4.66\textdegree&8.00\textdegree&6.02\textdegree\\ \hline
    DilatedNet \cite{chen2018appearance}&NO&PAR&4.42\textdegree&8.38\textdegree&6.19\textdegree\\ \hline 
    iTracker \cite{krafka2016eye}&NO&LR-EH&4.33\textdegree&7.12\textdegree&5.28\textdegree \\ \hline
    iTracker-MHSA \cite{cai2021gaze}&YES&LR-EH&4.05\textdegree&7.06\textdegree&5.17\textdegree \\ \hline
    GazeTR-Hybrid \cite{cheng2021gaze}&YES&Face Only&4.18\textdegree&7.12\textdegree&5.33\textdegree\\ \hline
    GazeCADSE \cite{o2022self}&YES&Face Only&4.04\textdegree&7.00\textdegree&5.25\textdegree\\ \hline
    \textbf{Proposed} &YES&EH-LR&\textbf{3.88}\textdegree& \textbf{6.46}\textdegree & \textbf{4.89}\textdegree \\ \hline
    \end{tabular}
    \label{tab:sota}
\end{table*}

Table \ref{tab:sota} shows the angular errors of each method on the evaluation datasets: MPIIFaceGaze, RT-GENE, and EYEDIAP. As iTracker and iTracker-MHSA did not provide the performance on the evaluation datasets, we re-implemented them by replacing their backbones with ResNet18 for fair comparison. In the table, among existing models, FullFace \cite{zhang2017s}, GazeTR \cite{cheng2021gaze}, and GazeCADSE \cite{o2022self} only use the full face image as the input for gaze estimation. RT-GENE \cite{fischer2018rt} feeds two cropped eyes to a VGG16 model. DilatedNet \cite{chen2018appearance} fuses the features of the left eye, right eye, and head directly.  iTracker \cite{krafka2016eye}, iTracker-MHSA \cite{cai2021gaze} fuse the features of the left and right eyes first then with the head features. Our proposed method also uses both the face and the eye images as inputs but has different ways of feature fusion we fuse the features of each eye and head in the first stage and then fuse the left and right features in the second stage. In addition, GazeTR, GazeCADSE, and our proposed methods utilize the transformers in the model. We show different types of gaze estimation models in Fig. \ref{fig:types}.

As shown in Table \ref{tab:sota}, our proposed methods TTGF with GAM achieved the state-of-the-art performance of gaze estimation on all the selected evaluation datasets. Among the methods using the feature fusing, our eye-head first then left-right combination shows the best performance. Overall the transformer-based methods show advantages in the performance of gaze estimation compared with non-transformer methods. Among the transformer-based methods, our model uses both the face and eye images, we used RoI alignment to resize the eye region to $128\times 128$, which enables the model to extract features directly from the eye patches. 

\begin{table}
    \centering
    \caption{Comparison of Computational Costs.}
    \begin{tabular}{l|c|c}
    \hline
    Model          & Params & FLOPs  \\ \hline
    RT-GENE \cite{fischer2018rt} & 82.0M & 30.81G  \\ \hline
    GazeTR-Hybrid \cite{cheng2021gaze} & 11.4M & 1.82G \\ \hline
    GazeCADSE \cite{o2022self}& 74.8M & 12.78G \\ \hline
    \textbf{proposed method} & 65.3M & 3.03G \\ \hline
    \end{tabular}
    \label{tab:param}
\end{table}

By using GAM, our proposed model achieves better performance on multiple datasets using only a single main model. This results in a smaller number of parameters compared with other methods. Suppose the number of parameters of the feature extractor is $N$ and that of each gaze regressor is $K$. For $M$ datasets, without GAM we need to train $M$ models for each dataset resulting in total $MN$ parameters. On the contrary, by applying GAM to train on multiple datasets, we only need one single model with one feature extractor, one gaze regressor and $M-1$ MLPs as the gaze offset for the anchor set is always $\boldsymbol{0}$. So the total number of parameters for our proposed model is $N + MK$. As $K$ is much smaller than $N$, our method needs fewer parameters to achieve better performance. 

Table \ref{tab:param} shows the number of parameters and the flops for each model. We can see that our proposed method has a fairly low computational cost which we believe is related to two reasons:  1) a relatively smaller model ResNet18 is applied as the backbone, and 2) a smaller size for the two eye patches as inputs.

\subsection{Ablation Study}
To study the individual contributions of the TTGF and GAM modules, we conducted ablation experiments by removing one of them from the entire framework. 

\begin{table*}
    \centering
    \caption{Ablation study.}
    \begin{tabular}{l|c|c|c|c}
    \hline
    Model          & Multiple Sets & MPIIFaceGaze & RT-GENE & EYEDIAP  \\ \hline
    itracker-MHSA \cite{cai2021gaze} & NO & 4.05 \textdegree& 7.06\textdegree&5.17\textdegree\\ \hline
    TTGF-Only & NO &3.98\textdegree& 6.89\textdegree & 5.11\textdegree \\ \hline
    TTGF-Only & YES & 4.12 \textdegree& 7.14 \textdegree & 5.20\textdegree \\ \hline
    proposed method (TTGF+GAM) & YES &\textbf{3.88}\textdegree& \textbf{6.46}\textdegree & \textbf{4.89}\textdegree \\ \hline
    \end{tabular}
    \label{tab:ablation}
\end{table*}

\subsubsection{Effect of TTGF}

To study the TTGF, we trained a TTGF-only model on each evaluation dataset and compared the results with itracker-MHSA. We compare with itracker-MHSA because it also uses a transformer encoder to combine eye and head features in a different order.
The itracker-MHSA fuses features first from the left and right eyes and then with the head feature. TTGF fuses features from each eye with head features and then across the two eyesAs we mentioned before, we re-implemented itracker-MHSA with the same backbone as our model for a fair comparison.
As we mentioned before, we re-implemented itracker-MHSA with the same backbone as our model for a fair comparison.

Table \ref{tab:ablation} shows the angular errors of each method on the evaluation datasets.  
The TTGF-only model outperforms the itracker-MHSA on all evaluation datasets. 

\subsubsection{Effect of GAM} 

We compared our proposed model with GAM with the TTGF-only model trained on multiple datasets. Table \ref{tab:ablation} shows that with GAM the accuracy of the TTGF-only model without multiple sets of training is improved on all three datasets from 0.1\textdegree to 0.43\textdegree respectively. 

To confirm the performance gain in multiple dataset training is due to the use of GAM, we trained the TTGF-only model with the combination of the ETH-XGaze and the evaluation datasets. The TTGF-only model trained on mixed datasets performed even worse than the TTGF-only model trained on each single evaluation set. This supports our claim that GAM can address the inconsistency in annotation across different datasets.

\section{Conclusion}
We proposed a Two-stage Transformer-based Gaze-future Fusion (TTGF) and the use of Gaze Adaption Modules (GAMs) for improving gaze estimation accuracy. The TTGF uses two-stage fusion for the features of the head and eye images through three transformer encoders. The proposed GAM generates gaze corrections to gaze estimates for one dataset (chosen here to be ETH-Gaze) to create estimates for images from other datasets. Our experiments show that our method surpasses the state-of-the-art by a significant margin. Ablation studies show that both innovations result in improvements when applied in isolation and that improvements compound when they are applied together. However, our proposed model still has some limitations. For example, the proposed TTGF needs cropped eye patches as input. The GAM does not address all issues arising from annotation inconsistency among gaze datasets. 

%
%
%
%

\end{document}